%% file: main.tex
\documentclass[
11pt, 
oneside, 
english, 
onehalfspacing, 
nolistspacing, 
headsepline, 
]{MastersDoctoralThesis} 

\usepackage[utf8]{inputenc} 
\usepackage[T1]{fontenc} 

\usepackage{mathpazo} 

\usepackage[backend=bibtex,style=numeric,natbib=true, maxnames=50, sorting=none]{biblatex} 

\usepackage[autostyle=true]{csquotes} 

\thesistitle{Active Perception with Neural Networks} 
\examiner{} 
\author{Elijah S. \textsc{Lee}} 

\AtBeginDocument{
\hypersetup{pdftitle=\ttitle} 
\hypersetup{pdfauthor=\authorname} 
}

\begin{document}
\hypersetup{linkcolor=black}

\frontmatter 

\pagestyle{plain} 


\begin{titlepage}
\begin{center}

\vspace*{.06\textheight}
\textsc{\Large WPE-II Written Report}\\[0.5cm] 

\HRule \\[0.4cm] 
{\huge \bfseries \ttitle\par}\vspace{0.4cm} 
\HRule \\[1.5cm] 
 
\emph{Author:}\\
\href{https://sites.google.com/view/elijahslee/}{\color{black}{\authorname}} 
\vfill
Department of Computer and Information Science\\
University of Pennsylvania\\
Philadelphia, PA 19104
\vfill
\vfill
\end{center}
\end{titlepage}

\addchap*{Abstract}
\addchaptertocentry{Abstract} 
Active perception has been employed in many domains, particularly in the field of robotics. The idea of active perception is to utilize the input data to predict the next action that can help robots to improve their performance. The main challenge lies in understanding the input data to be coupled with the action, and gathering meaningful information of the environment in an efficient way is necessary and desired. With recent developments of neural networks, interpreting the perceived data has become possible at the semantic level, and real-time interpretation based on deep learning has enabled the efficient closing of the perception-action loop. This report highlights recent progress in employing active perception based on neural networks for single and multi-agent systems.

\tableofcontents 

\mainmatter 

\pagestyle{thesis} 


\include{Chapters/chapter1}

\include{Chapters/Chapter2} 
\include{Chapters/Chapter3}
\include{Chapters/Chapter4} 
\include{Chapters/Chapter5} 






\printbibliography[heading=bibintoc]


\end{document}

%% file: Chapters/chapter1.tex
\chapter{Introduction} 
\label{chapter1} 

About 30 years ago, Bajcsy introduced the concept of active perception \cite{bajcsy1988active}. The main idea of active perception is to reposition sensors to gather information useful for better understanding the world. Compared to a static agent that passively senses its environments and takes an action based on a given perception, an active agent can improve the quality of perception reasoning by actively moving around the surroundings and thus gather more meaningful information to improve its performance.  

\begin{figure}[h]
\centering
\includegraphics[width=.9\textwidth]{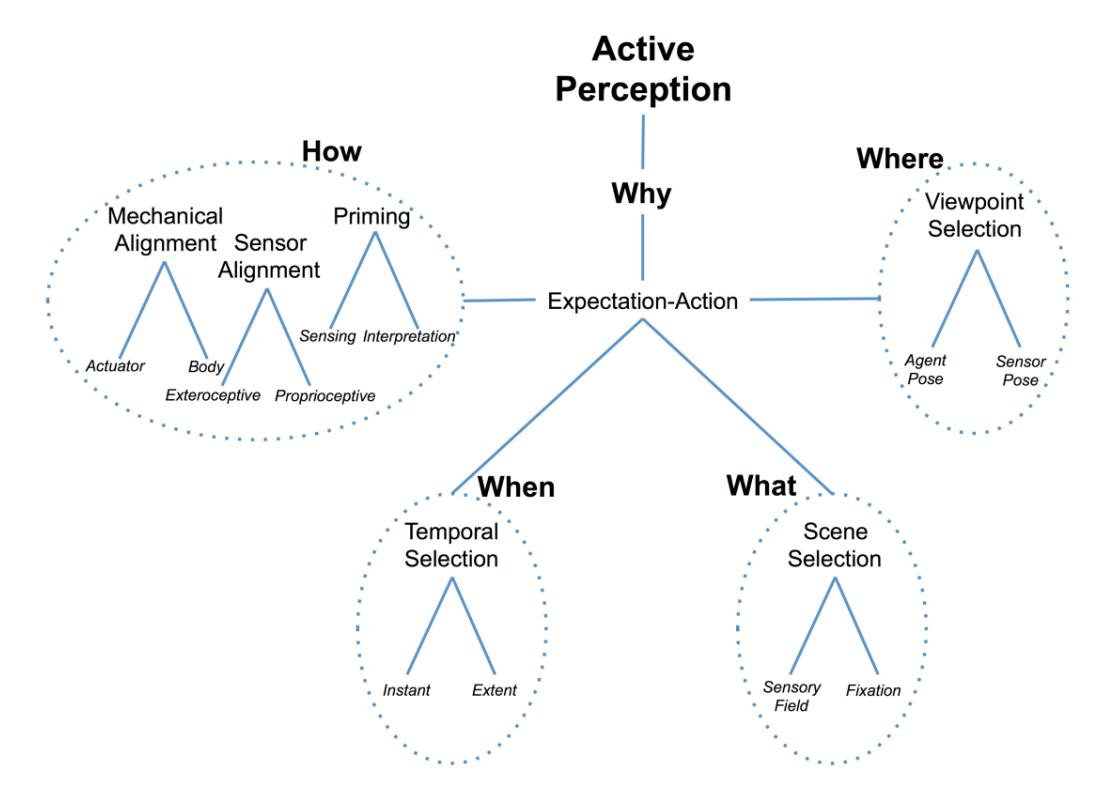}
\caption{The elements of Active Perception from \cite{bajcsy2018revisiting}.}
\label{fig:active_tree}
\end{figure}

Figure \ref{fig:active_tree} presents detailed elements of the active perception, which is adapted from Bajcsy \cite{bajcsy2018revisiting}. 
%
%
In this figure, the author explains the concept of active perception by listing various motivations. For instance, one aspect of active perception could be moving the sensor itself (e.g. changing actuator or sensor alignment under \textit{How} component) to gather more information about the world. Another example can be a framework that prioritizes certain sensory modalities for the information gain (e.g. scene selection under \textit{What}, temporal selection under \textit{When}, and viewpoint selection under \textit{Where} components). Scene selection may occur when a robot only uses an ROI of image data for perception. Using information gain from a specific time frame for better perception is temporal selection, and choosing different viewpoints results in viewpoint selection. In this sense, active perception is distinct from passive (single or multi-view) perception because the essence of active perception involves either moving sensor or choosing which sensor and which portion of the sensor data one selects. This report explores different approaches in executing active perception and shows how these methods outperform passive perception. The performance metric is determined based on the task (e.g. MSE error for object tracking or Average Precision for bounding box detection). 
%

Over the past few years, there have been intensive research activities employing neural networks. Perception systems based on deep learning with neural nets are increasingly used in everyday life. End-to-end approaches using deep neural networks have been proven tremendously successful in many computer vision tasks such as object detection, semantic segmentation, and pose estimation. In this report, we investigate how deep learning can help active perception. 

Traditional methods without deep learning employed active perception in settings where perception is simplified. In Gürcüoglu et al. \cite{gurcuoglu2013hierarchical}, a 3D sensor is utilized to detect the small cube target for active perception; however, a range sensor cannot provide semantic information so this work presumes prior knowledge of the target shape. In \cite{falanga2017aggressive}, the aggressive quadrotor uses an onboard camera to detect the narrow gap with a black-and-white rectangular pattern, which simplifies the perception task with image processing. In \cite{acevedo2020dynamic}, it is assumed that the robots are equipped with Radio Frequency Identification (RFID) readers to search for a child wearing an RFID tag. Other methods perform simplified perception using fiducial markers \cite{olson2011apriltag}.

In order to perform active perception tasks without simplified perception, recent works utilize deep learning based on convolutional neural networks. Neural networks are known to learn the nonlinear relationship between perception and necessary information so a full active perception system is possible and closes the perception-action loop. Drones are actively moved to optimize the viewpoint for deep learning-based human pose estimation \cite{kiciroglu2020activemocap}. A camera is moved sequentially to get a better estimate of the pose by leveraging neural nets and active pose estimation \cite{ren2019domain}. An unmanned aerial vehicle employs active perception by executing a pre-trained off-the-shelf deep learning-based object detector in searching for a lost child \cite{sandino2020autonomous}. Deep reinforcement learning is performed when a robotic manipulator interacts with a scene for improved object finding in clutter \cite{novkovic2020object}. The survey in this field is detailed in \cite{gallos2019active}.

This report explores active perception techniques within the context of neural networks. Based on three representative works \cite{ren2019domain,price2018deep,wang2020v2vnet}, we discuss how neural networks help active perception with (i) vision-based single robot \cite{ren2019domain}; (ii) vision-based multiple robots \cite{price2018deep}; and (iii) lidar-based multiple vehicles \cite{wang2020v2vnet}. Each of these works also represents an element of active perception shown in Figure \ref{fig:active_tree}: viewpoint selection \cite{ren2019domain}, scene selection \cite{price2018deep}, and temporal selection \cite{wang2020v2vnet}.



%% file: Chapters/chapter2.tex
\chapter{Active Perception with Vision-based Single Agent} 
\label{chapter2} 
Ren et. al \cite{ren2019domain} employ active perception with vision-based single robot. In particular, they tackle solving a state estimation problem that is relevant to identifying the positions of objects in the scene, forming the basis of the manipulation task. The main idea is to move the robot to get a better estimate of an object pose, and a camera is used as the perception sensor. This work performs viewpoint selection of active perception by changing agent pose and sensor pose.

\section{Overview}
This work aims to improve the accuracy of pose estimation by making the observation that robots interact with objects in the scene. In the setting, the geometrical 3D model of an object $x$ is assumed with some reference object $y$. Let $O_y$ denote a coordinate frame relative to $y$, and let $P_x$ denote
the 6D pose of $x$ in the coordinate frame $O_y$. The goal is to estimate $P_x$ from the image $I$ that contains $x$ and $y$.

\begin{figure}[h]
\centering
\includegraphics[width=.7\textwidth]{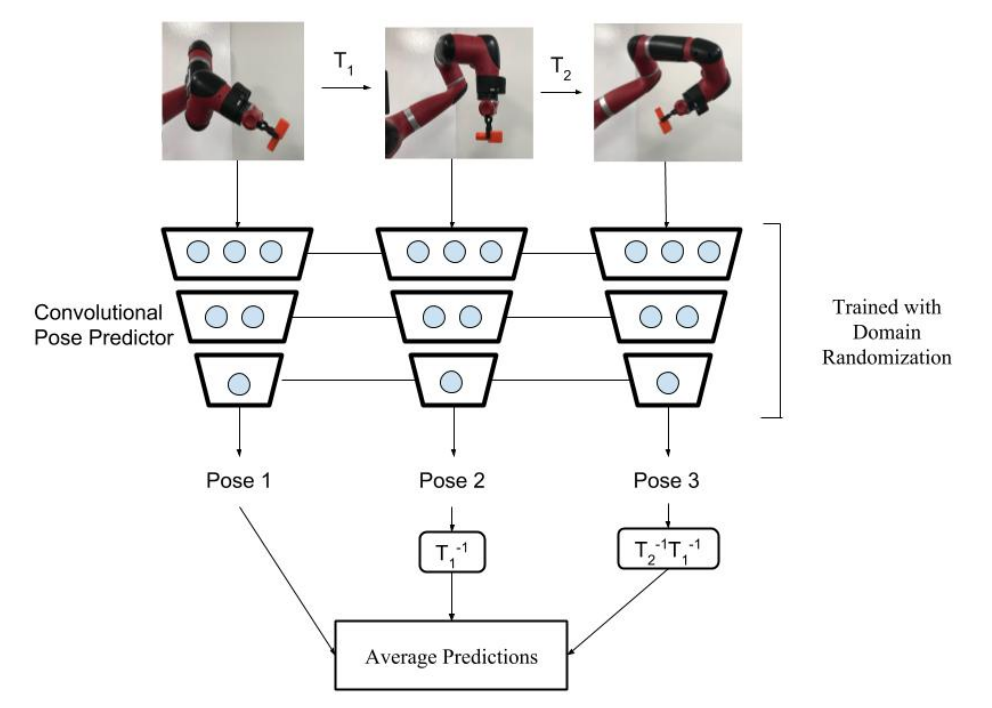}
\caption{Overview of single-agent active pose estimation.}
\label{fig:overview}
\end{figure}

Figure \ref{fig:overview} shows the overview of the proposed approach. The given image $I$ and its known transformations through $T_1$ and $T_2$ are input to the convolutional neural networks trained with domain randomization \cite{tobin2017domain}, and the final pose is improved by combining the neural net outputs. 

\subsection{Neural Networks for Pose Estimation}
Pose estimation has been studied extensively based on 2D model \cite{collet2011moped}, 3D model \cite{ye2011accurate}, and fiducial markers \cite{olson2011apriltag}. Since these approaches depend on features or markers, methods based on supervised learning with neural networks \cite{tekin2018real} has been explored. PoseCNN \cite{xiang2017posecnn} trains the network with a translation of an object center and rotation as a quaternion representation. DenseFusion \cite{wang2019densefusion} estimates the pose from RGB-D images integrating an end-to-end iterative pose refinement procedure. MSL-RAPTOR \cite{ramtoula2020msl} combines learning with a tracker for onboard robotic perception. To avoid extensive labeling in supervised learning methods, domain randomization \cite{tobin2017domain} generates its own training data in simulation, which learns a robust policy that transfers well to the real world. This work combines domain randomization with active perception to improve the accuracy of pose estimation of an object.

\subsection{Active Perception based on Domain Randomization}
\begin{figure}[h]
\centering
\includegraphics[width=1\textwidth]{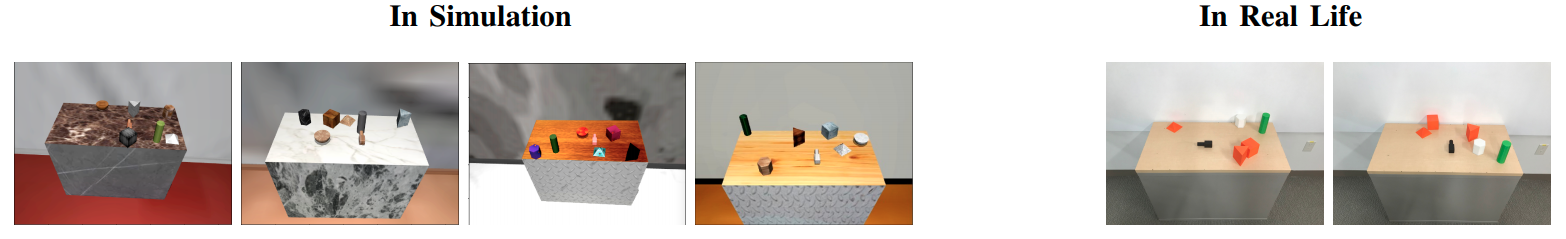}
\caption{Sim2real transfer via domain randomization.}
\label{fig:domain}
\end{figure}

Domain randomization \cite{tobin2017domain} is known to work well in bridging the gap between simulation and the real world by varying visual properties in the scene. In training neural nets, this method helps the networks learn a robust policy that transfers well from simulation to the real world and avoids extensive labeling in supervised learning. As displayed in Figure \ref{fig:domain}, various visual domain in simulation is used in collecting the training set, and simulation to reality transfer is achieved in testing the trained networks.

\textbf{Domain Randomization with Geometric Transformations} problem (DR-GT): Let $T_1,...,T_k$ denote a set of $k$ transformations that are actively applied to the scene. The proposed training data in the form of tuples are

\begin{equation}
\{I,T_1(I),...,T_k(I),P_x,T_1(P_x),...,T_k(P_x)\}    
\end{equation}

Then, the authors hope to learn a mapping from $I,T_1(I),...,T_k(I),T_1,...,T_k$ to $P_x$. Let $f$ be the domain randomization mapping from $I$ to $P_x$ and let $T_i^{-1}$ denote the inverse transform of $T_i$. Then, they compute a series of predictions by
\begin{equation}
P_{x;0} = f(I), \hspace{3mm}
P_{x;1} = T_1^{-1}(f(T_1(I))), \hspace{3mm}
...\hspace{3mm}, \hspace{3mm}
P_{x;k} = T_k^{-1}(f(T_k(I)))
\end{equation}
where $P_{x;i}$ is an estimate of $P_x$. The authors predict $P_x$ as the sample average:

\begin{equation}
    \hat{P_x} = \frac{1}{k+1}\sum_{i=0}^k P_{x;i}
\end{equation}

\textbf{Model Architecture:} The proposed approach predicts 3DoF pose composed of 2DoF translation and 1DoF rotation. The neural networks take in an RGB camera image through 16 convolutional layers, each of two layers is followed by a max-pooling and ReLU functions. The convolutional layers are followed by 3 fully connected layers. The architecture is similar to VGG, using pre-trained networks on ImageNet. The training loss function consists of L1 regression loss for 2DoF translation and a cosine loss for the 1DoF rotation, defined as

\begin{equation}
    L(x,\theta) = ||x-\hat{x}|| + || \cos(\theta-\hat{\theta})-1 ||
\end{equation}

\textbf{Performance Evaluation:} This work evaluates the algorithm performance in three scenarios: (1) Moving reference objects in the environments, (2) Moving a robot manipulator holding an object, and (3) Moving a camera held by the robot. In all cases, adding additional viewpoints improves the pose estimation results. The 1.5cm pose estimation error in domain randomization state-of-the-art is reduced to less than 0.6cm.

\section{Evaluation}
In this section, we discuss active perception based on neural networks with a single agent. This work performs viewpoint selection of active perception elements. The method for active pose estimation exploits the fact that being able to see an object from different angles and in different positions leads to a more accurate and robust prediction. In executing active perception, the agent moves around the object, and also a manipulator holding an object can move; thus, a single agent can perform active perception by applying a geometric transformation between the robot and the object. Domain randomization enables sym2real transfer without much labeling on the training data and relatively simple neural networks are used as the loss function. The performance metric is an average error for position and orientation of the object pose, and the active perception result outperforms passive perception, which utilizes a single viewpoint.

Some may argue that it is ambiguous to claim that this work incorporates active perception since this work can be viewed as a multi-view perception rather than an active perception. This argument foreshadows potential extensions of this work if agents apply forceful interactions for better perception - for instance, a robotic manipulator might interact with a scene to improve pose estimation of an object in clutter, which opens up a problem in interactive perception \cite{bohg2017interactive}. 

A weakness of this approach is that a single agent has to move around the scene to allow various viewpoints, which requires more dynamics and time for the agent to collect the information of the scene.
Applying the idea of this work to multiple agents would be a reasonable future extension. This work only utilizes multiple images for active perception, so it would be also better to use more images to strengthen the results, and a multi-agent system may provide multiple data points easily. Further, the proposed method for active perception based on domain randomization assumes no error in transformation among different viewpoints. In real-world applications, noise and uncertainty have to be considered for robust algorithm performance, which offers another potential extensions. 

%% file: Chapters/chapter3.tex
\chapter{Active Perception with Vision-based Multiple Agents} 
\label{chapter3} 
Price et. al \cite{price2018deep} tackle deep neural network-based visual tracking based on active perception with vision-based multiple aerial vehicles. They run a team of cooperating micro aerial vehicles (MAVs) with on-board cameras to achieve on-board, real-time, continuous, and accurate vision-based detection and tracking of a person. Their approach maintains an active perception-driven formation of scene selection that exploits the power of DNNs for visual object detection.

\section{Overview}
This work aims to measure the position and velocity of the tracked person in the world frame $W$ at time $t$, denoted by $^W x_t^P \in \mathbb{R}^6$. The uncertainty covariance matrix associated to it is given by $^W \Sigma_t^{P} \in \mathbb{R}^{6\times 6}$. Let there be $N$ MAVs $R_1,...,R_N$ tracking a person $P$. Let the 6D pose of the $n_{th}$ MAV in the world frame at time $t$ be given by $^W x_t^{R_n} \in \mathbb{R}^6$. The uncertainty covariance matrix associated to the MAV pose is given by $^W \Sigma_t^{R_n} \in \mathbb{R}^{6\times 6}$. Each MAV $n$ has an on-board camera $C_n$.

\subsection{Neural Networks for Target Tracking}
Target tracking has been extensively studied in the past years. Traditional methods had exploited feature-based detection but the recent development of neural networks has enabled agents to fully understand semantic information, which results in deep learning-based detectors \cite{schmidhuber2015deep}. However, there are only a few works that exploit the power of deep neural networks for visual object detection onboard MAVs. The main bottleneck is the computational requirements since the computing power of a dedicated high-end GPU is required for real-time detection but such hardware is usually bulky and thus impractical to install onboard a MAV. Some networks suitable for real-time detection include YOLO \cite{redmon2016you} or Faster R-CNN \cite{ren2015faster}, which are both outperformed in speed and detection accuracy by SSD multibox approach \cite{liu2016ssd}. This work uses SSD multibox detector in the scheme of active perception involving an active selection of the most informative ROI.

\subsection{Active Perception for Cooperative Detector and Tracker}
\begin{figure}[h]
\centering
\includegraphics[width=0.99\textwidth]{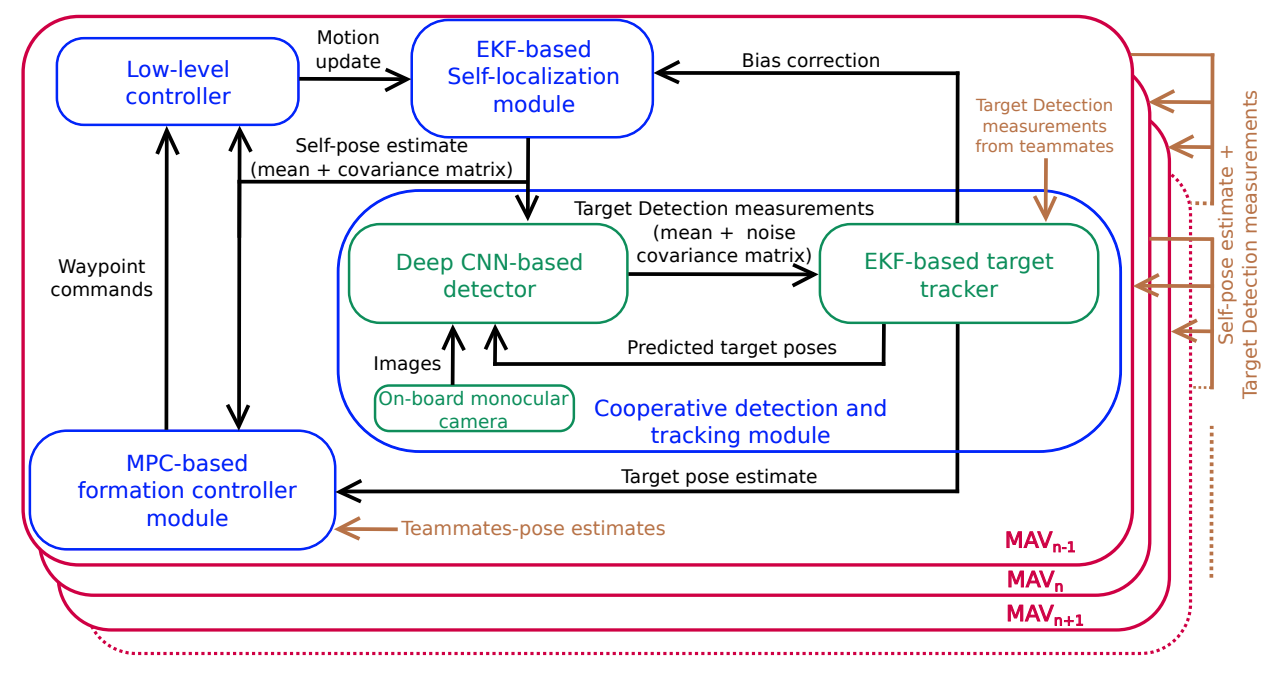}
\caption{Overall architecture of cooperative tracking system.}
\label{fig:target_overview}
\end{figure}

Although SSD is relatively efficient, real-time onboard processing of high-resolution images is still infeasible. A naive approach is to process a down-sampled low-resolution image but this will be sub-optimal due to the reduced information. Accordingly, DNNs would often fail at objects on small scale or far away from the camera, which is a typical scenario with aerial vehicles. To remedy this, this work leverages the mutual world knowledge that is jointly acquired by multi-robot systems and shares compactly representable detection outputs with low bandwidth requirements. As a result, each MAV knows where to head in the future and performs active perception by selecting the relevant ROI that provides the highest information content. Figure \ref{fig:target_overview} details the flow of data in the overall architecture and shows that MAVs share data of their self-pose estimates and the detection measurements of the tracked target.

\newpage

\begin{figure}[h]
\centering
\includegraphics[width=0.7\textwidth]{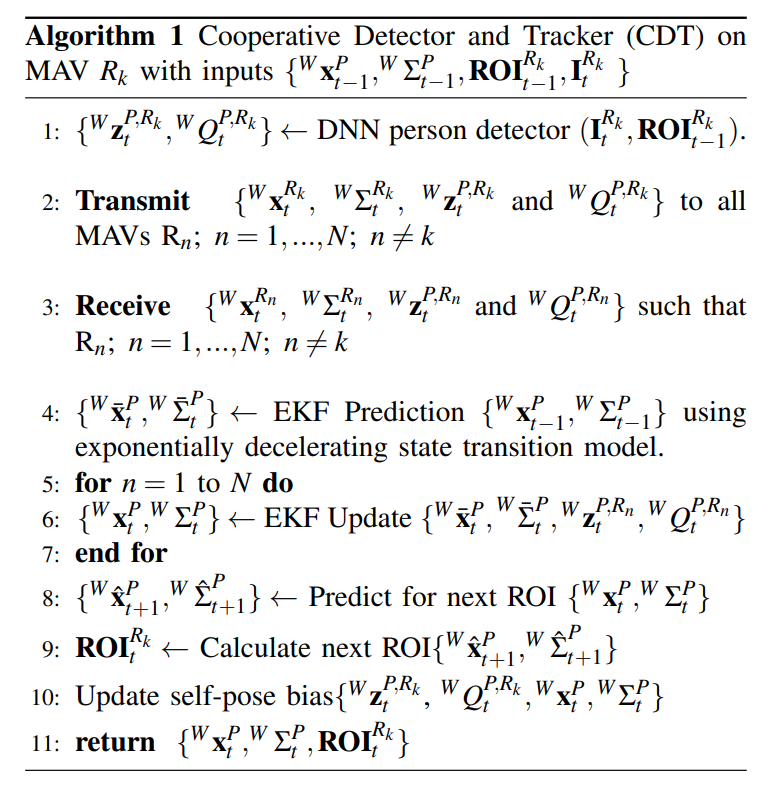}
\caption{Algorithm for cooperative detection and tracking.}
\label{fig:target_algorithm}
\end{figure}

\textbf{Target Tracking Algorithm: } Figure \ref{fig:target_algorithm} demonstrates the overall algorithm. The system starts with detecting a person using deep neural networks on an ROI from the previous time step $t-1$ of the current image. The detection measurements, computed as a mean $^W z^{P,R_k}$ and a noise covariance $^W Q_t^{P,R_k}$, are transmitted and received in Step 2 and 3. Step 4 performs the prediction of the EKF and Steps 5-7 combine measurements from all MAVs with the EKF update. In Steps 8-9, the algorithm actively selects an ROI ensuring that future detections are performed on the most informative part of the image while keeping the computational complexity independent of camera image resolution. Step 10 performs the bias update of the MAV self-pose, and Step 11 returns the updated mean and covariance of the person with ROI.

In predicting the state of the person in the next timestamp, the algorithm calculates the position and associated uncertainty of the person's head and feet. To do so, the height distribution model for a person is assumed, and the person is assumed to be in an upright position. The head and feet positions are then back-projected onto the image frame along with the uncertainties, and the left and right borders of the ROI are calculated to match the desired aspect ratio (e.g. 4:3).

\textbf{Model Architecture:} The proposed approach uses pre-trained SSD Multibox detector \cite{liu2016ssd} for locating a person. The output of the detector is a bounding box with a class label and a confidence score. The input ROIs of images are down-sampled to 300 by 300 pixels. Detection is projected into 3D, shared among MAVs, and fused using an Extended Kalman Filter on each MAV. After correcting the self-pose bias, the MAVs maneuver in a formation using an MPC-based controller maintaining (i) a preset distance to the tracked person, (ii) a preset altitude above the ground plane, and (iii) orientation towards the tracked person. A potential-field-based collision avoidance system is implemented on top of MPC.

\textbf{Performance Evaluation:} This work evaluates the algorithm performance in two scenarios: (1) stationary person experiment and (2) moving person experiment. The following datasets are generated to compare the results:

\begin{itemize}
    \item No MAV self-pose bias correction (\textbf{No SPBC}): The MAV self-pose bias correction is not used.
    \item No actively-selected ROI (\textbf{No AS-ROI}): Active ROI selection is not used.
    \item \textbf{MAV 1 only} or \textbf{MAV 2 only}: Single MAV is used with active ROI selection. 
\end{itemize}

\begin{figure}[h]
\centering
\includegraphics[width=0.4\textwidth]{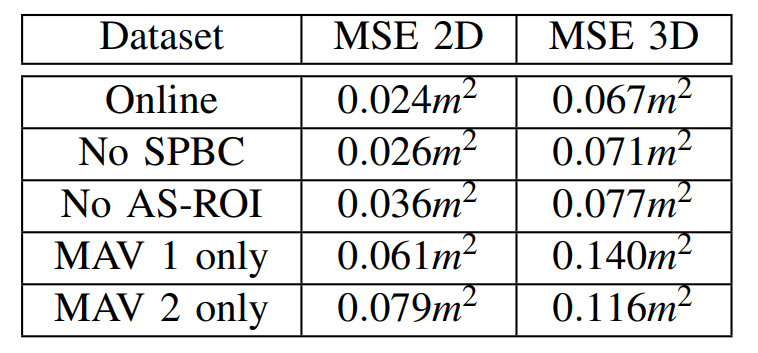}
\caption{Stationary person experimental result.}
\label{fig:target_tab1}
\end{figure}

\begin{figure}[h]
\centering
\includegraphics[width=0.72\textwidth]{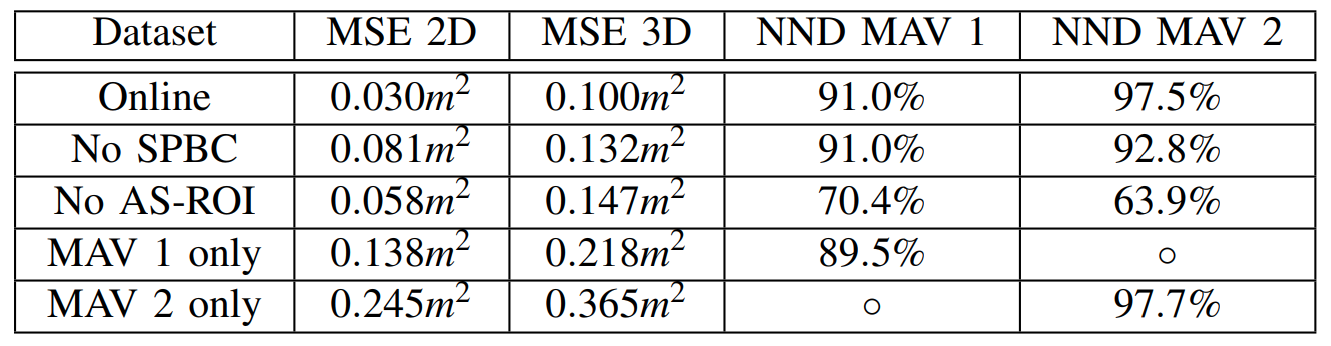}
\caption{Moving person experimental result.}
\label{fig:target_tab2}
\end{figure}

Figure \ref{fig:target_tab1} and Figure \ref{fig:target_tab2} show the experimental results for stationary person and moving person, respectively. The benefit of active ROI selection is more significant in the dynamic situation, and overall cooperative perception outperforms single MAV perception in both cases.

\section{Evaluation}
We discuss active perception based on neural networks with a single agent. This work achieves real-time, continuous, and accurate DNN-based tracking with active perception-based MAVs. In an active perception scheme, this method belongs to the category of scene selection, which selects the relevant sensory field for better perception. Although the computational complexity of DNN-based detectors may grow very fast with the image resolution, the scene selection approach actively selects ROIs for the DNN-based detectors, which enables real-time detection with high accuracy. Another strength of this paper is the full-stack architecture of the robotic platform where the authors combine planning, perception, and control modules all together to simulate MAV-based target tracking of humans. In this sense, the active perception is successfully coupled with other modules, and the lower MSE error of the human pose verify the novelty of the proposed algorithm with active ROI selection. Moreover, the cooperative detection and tracking method accounts for a realistic noise model of the person detector with noise quantification. 

A weakness of this approach is that the detector is constrained to detect a person, and thus the algorithm performance may be largely dependent on the dataset. This work assumes that the person is standing or walking upright in the world frame, so the performance may be biased to the dataset. The number of MAVs is limited to two during the experiment. Diversifying the dataset will improve the performance, and combining scene selection from more than two agents could be a possible extension.

%% file: Chapters/chapter4.tex
\chapter{Active Perception with Lidar-based Multiple Agents} 
\label{chapter4} 
Wang et. al \cite{wang2020v2vnet} implements V2VNet, which performs active perception with lidar-based multiple vehicles. Particularly, they explore the use of vehicle-to-vehicle (V2V) communication with lidar sensors to improve the perception and motion forecasting performance of autonomous vehicles. They argue that observing the same scene from different viewpoints allows them to see through occlusions and detect actors at long range. Their work also shows that sending compressed deep feature map activations achieves high accuracy while satisfying communication bandwidth requirements. This approach pursues temporal selection of active perception in that the algorithm selects information from different vehicles coming from the V2V communication. 

\section{Overview}
\begin{figure}[h]
\centering
\includegraphics[width=0.8\textwidth]{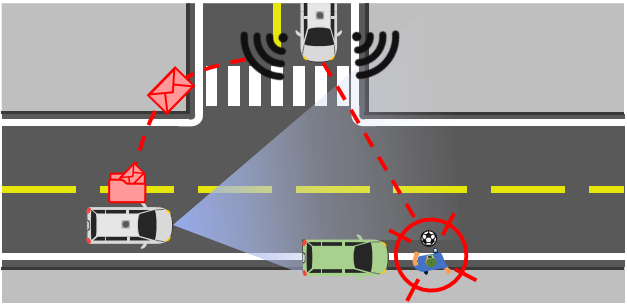}
\caption{Safety critical scenario of a pedestrian coming out of occlusion.}
\label{fig:v2v_intro}
\end{figure}

A core component of autonomous vehicles is to perceive the world and forecast how other agents will maneuver. In this section, the authors tackle the perception and motion forecasting problem and discuss how active perception within the context of the V2V communication setting improves the algorithm performance. Figure \ref{fig:v2v_intro} shows a safety-critical scenario of a pedestrian coming out of occlusion where V2V communication can be leveraged to see the scene from different viewpoints. 

In V2V perception, the receiving vehicle should aggregate information from different viewpoints such that its field of view is maximized. As active perception focuses on deciding which action the agent should take to better perceive the world, V2VNet considers multiple autonomous vehicles to see the environments better by sharing perception messages.   

\subsection{Neural Networks for Perception and Motion Forecasting}
From sensor data, the autonomous vehicles need to reason about the scene in 3D, identify other agents, and forecast their future maneuvers. Thus, perception and motion forecasting are critical for self-driving vehicles to plan through traffic to get from one point to another safely. The robustness of perception and motion forecasting algorithms has been significantly improved in the past years due to the advancement of neural networks. Liang et al. \cite{Liang_2019_CVPR} fuse multi-sensor data with neural networks for 3D object detection. Casas et al. \cite{casas2020implicit} perform motion forecasting by modeling the scene as an interaction graph and employing graph neural networks to learn a distributed latent representation of the scene. Recently, Luo et al. \cite{luo2018fast} propose a novel deep neural network that is able to jointly reason about 3D detection, tracking, and motion forecasting given data captured by a 3D sensor, dubbed perception and prediction (P\&P). Despite all these advances, challenges remain when objects are heavily occluded or far away. The main bottleneck is that these works only employ a single viewpoint. The V2V communication for active perception combines information from different viewpoints, and this work utilizes a spatially-aware graph neural network (GNN) to aggregate the information received from all the nearby agents.

\subsection{Active Perception within the context of V2V setting}
There are two key questions in the V2V setting: (i) what information should each agent transmit? (ii) how should each agent incorporate the information received from other agents for optimal perception and motion forecasting? V2VNet effectively answers these questions by employing active perception. Figure \ref{fig:overview} shows the proposed V2VNet architecture. 

\begin{figure}[h]
\centering
\includegraphics[width=1\textwidth]{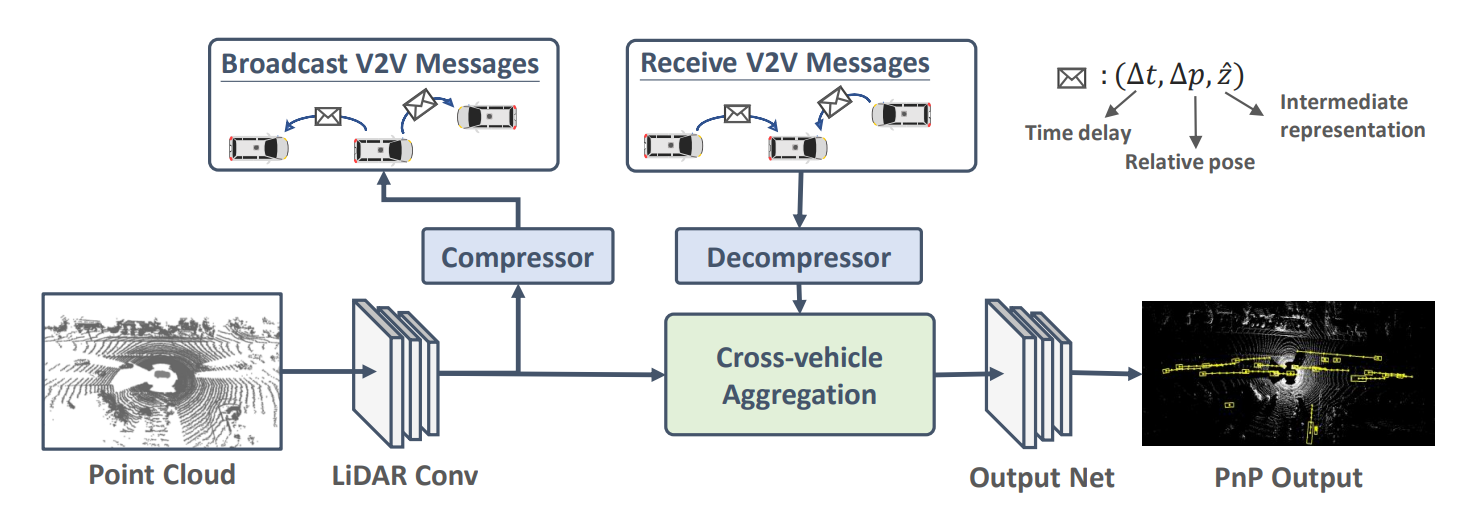}
\caption{Overview of V2VNet.}
\label{fig:domain}
\end{figure}

\textbf{Three types of information for broadcasting:} An agent can choose to transmit three types of information: (i) the raw sensor data, (ii) the intermediate representations of perception and prediction system, or (iii) the output detections and motion forecast trajectories. Although all three types are valuable, the authors argue that there are tradeoffs among these types because they want to minimize the message size but increase the data accuracy. In the active perception scheme, this work argues that sending intermediate representations of the perception and prediction network achieves the best performance. Furthermore, it is shown in \cite{wei2019learning} that intermediate representations in deep networks can be easily compressed while retaining important information for downstream tasks.   

\textbf{Model Architecture:} As Figure \ref{fig:overview} demonstrates, V2VNet has three main blocks: (i) a convolutional neural network that processes raw point cloud data, (ii) a cross-vehicle aggregation module, and (iii) an output network that computes the final P\&P outputs. 

In the LiDAR convolution block, the authors voxelize the past five LiDAR point cloud sweeps and apply 3 by 3 convolution filters (with strides of 2, 1, 2) to produce a 4x downsampled spatial feature map. The resulting output is the intermediate representation.

The aggregation module takes in messages from different spatial locations at different timestamps. In this module, each vehicle uses a fully connected graph neural network (GNN), where each node in the GNN is the state representation of an autonomous agent in the scene. GNN is a natural choice for the neural network since it handles dynamic graph topologies, which arise in the V2V setting.

The output network is a set of four Inception-like \cite{szegedy2015going} convolutional blocks. The network takes in the feature map and outputs $(x,y,w,h, \theta)$, denoting the position, size, and orientation of each agent.

\begin{figure}[h]
\centering
\includegraphics[width=0.7\textwidth]{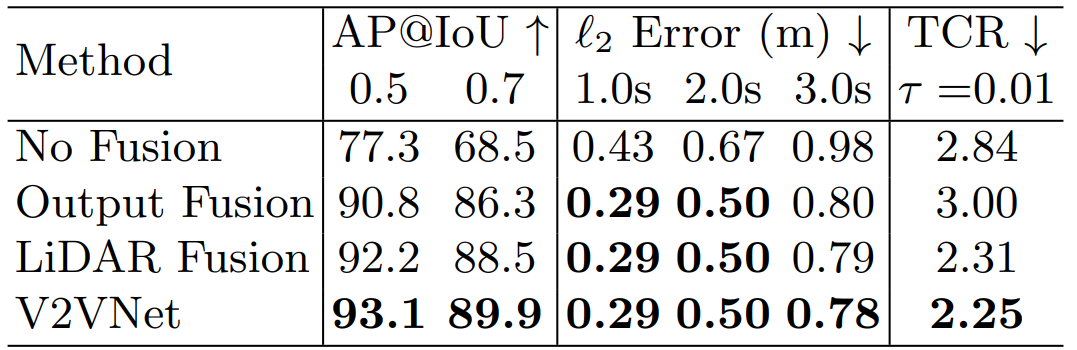}
\caption{Comparison for algorithm performance.}
\label{fig:v2v_result}
\end{figure}

\textbf{Performance Evaluation:} The authors evaluate the algorithm performance as shown in Figure \ref{fig:v2v_result}. The baseline method consists of LiDAR backbone network and output headers only, dubbed \textit{No Fusion}. For \textit{Output Fusion}, each agent sends post-processed outputs, i.e., bounding boxes with confidence scores, and \textit{LiDAR Fusion} warps all received LiDAR sweeps from other agents to the coordinate frame of the receiver and performs direct aggregation. As shown in Figure \ref{fig:v2v_result}, V2VNet models outperform all other baseline methods in both detection and prediction. V2VNet's performance gain over \textit{LiDAR Fusion} may come from utilizing GNN in the cross-vehicle aggregation stage to reason about different agents' feature maps more intelligently than the naive aggregation method.

\section{Evaluation}
In this section, we discuss active perception with Lidar-based multiple agents. This work utilizes various neural networks including GNN to improve the algorithm performance and efficiently process dense lidar data. Active perception of which sensor measurements are relevant to the ego vehicle is achieved via this network. In particular, temporal selection of information gain is performed by actively selecting the time-stamped data coming from V2V communication. This active perception method for perception and prediction achieves the best compromise between accuracy improvements and bandwidth requirements. Further, the collaboration among multiple agents allows agents to see through occlusions and detect actors at a long-range, where the observations are very sparse.

A limitation is that all vehicles are assumed to be localized in aggregating the messages from autonomous vehicles. Therefore, the active perception does not compensate for noise or error in transferring the V2V messages so the aggregation process may be erroneous. Incorporating the noise and uncertainty by formulating related covariance for vehicle pose would be a possible extension. Another limitation is that it is not clear whether the used dataset captures diverse scenarios in terms of vehicle positions and arrangements. The GNN architecture is dependent on the spatial information of V2V communication data, so variation in the dataset will strengthen the algorithm and can be generalized. 

%% file: Chapters/chapter5.tex
\chapter{Conclusions} 
\label{chapter5} 
In this section, we compare the presented approaches to neural network-based active perception described in previous sections. We also address possible future directions.    

\section{Comparison}
The three methods all use active perception within the context of neural networks, with a different focus on the task. Table \ref{tab:ap} summarizes the aforementioned approaches with respect to the active perception. All the works are components of active perception in that the agent can select viewpoint, scene, or time to improve overall perception. For the task, \cite{ren2019domain} solves pose estimation problem, \cite{price2018deep} tackles detection and tracking problem, and \cite{wang2020v2vnet} focuses on perception and motion forecasting. The used performance metrics are shown for each task.

\begin{table}[b]
\centering
\caption{Comparison of active perception.}
\label{tab:ap}
\begin{center}
\begin{tabular}{ |c|c|c| } 
 \hline
 \textbf{Active perception component} & \textbf{Task} & \textbf{Metric}\\
 \hline
 Viewpoint (Where) \cite{ren2019domain} & Pose estimation & Mean prediction error\\ 
 \hline
 Scene (What) \cite{price2018deep} & Object tracking & Mean square error \\ 
 \hline
 Temporal (When) \cite{wang2020v2vnet} & Motion forecasting & Average precision \\ 
 \hline
\end{tabular}
\end{center}
\end{table}

\textbf{Number of agents:} In large, \cite{ren2019domain} utilizes a single agent while \cite{price2018deep} and \cite{wang2020v2vnet} use multiple agents. Since the single agent needs to move around the scene to allow various viewpoints for active perception, the advantage of employing a multi-robot system is to reduce the dynamics and time required for the agents to collect the information of the scene. 

\textbf{Type of Sensor:} In perceiving the scene, \cite{ren2019domain} and \cite{price2018deep} mainly employ vision sensor for active perception while \cite{wang2020v2vnet} uses LiDAR sensor. It is a bit challenging to argue which sensor outperforms the other because any sensor has pros and cons. For instance, the camera is good for extracting semantic information and cost-effective while LiDAR provides accurate range measurements. The presented works focus on a single sensor but there may be a possible future direction in fusing various sensors for active perception.

\textbf{Neural Networks:} In terms of neural networks, \cite{ren2019domain} uses domain randomization technique for seamless sim2real transfer. The benefit of this method is that it does not require real-world data set because the environmental variation in a photo-realistic simulator allows the networks to learn features relevant to the real-world testing without much labeling. Relatively simple architecture similar to VGG is used for model architecture. \cite{price2018deep} uses a pre-trained SSD Multibox detector for locating a human. The novelty of this work is reducing the computational complexity of DNN-based detectors by actively selecting ROIs so that the approach is real-time with high detection accuracy. This work also accounts for a realistic noise and uncertainty that the other works do not consider. The V2VNet \cite{wang2020v2vnet} consists of three neural network blocks for point cloud processing, cross-vehicle aggregation module, and final output computation. Since LiDAR point cloud is distinct from the data of the previous two works, voxelizing the point cloud is needed to produce a spatial feature map, and the tradeoffs are considered among the three types of information: (i) the raw sensor data, (ii) the intermediate representations, or (iii) the output detections and motion forecast trajectories. Furthermore, GNN is utilized for the aggregation module to represent the agents as nodes in a fully connected graph.

In summary, we have explored various active perception techniques with neural networks. Active perception is distinct from passive single or multi-view perception in that it involves moving sensor or prioritizing sensory modalities to better understand the world. The three representative works demonstrate the various active perception components by viewpoint, scene, and temporal selection. We observe that viewpoint selection may be better suitable for multi-agent system where multiple agents can collaborate. We find that scene selection reduces algorithm computation by selecting and processing only portion of a scene. Temporal selection is necessary when multiple data with different time-stamp are transmitted to a single agent. Each component of active perception is a powerful tool for better perception, and thus using appropriate active perception technique can benefit the agent perception; furthermore, combining various components may improve the overall performance and offer interesting future directions.

\section{Future Directions}
There could be many possible future directions combining the advantages of the aforementioned approaches as well as considering various assumptions. We address a few directions as follows:

\textbf{Sensor Fusion:} Employing heterogeneous teams of robots with different sensors would enable sensor fusion-based active perception. If the camera provides semantic information while LiDAR gives dense point clouds, agents may find the perception very informative. How to aggregate the information from various sensors and what to expect for various viewpoints and perception data could be an interesting problem.

\textbf{Noise Modeling:} As \cite{ren2019domain} and \cite{wang2020v2vnet} demonstrate, the majority of the works based on active perception assume the perfect state estimation for the perceiving agents. As the agents can be dynamic, the transferred data between agents may not be perfect, and different poses of ego vehicles would result in inaccurate perception. Modeling the associated noise and uncertainty is interesting to investigate.

\textbf{Reinforcement Learning:} With recent developments of sim-to-real transfer, reinforcement learning can be adapted to a wide variety of applications and scenarios with good performance. Its simulation environments converge towards working control policies and stable inference. Employing active perception on a simulator with a reinforcement learning-based platform is promising and can be applied to real-world testing without extensive labeling. For instance, we can think of the setting in perimeter defense. In this game, the goal of intruders is to reach the perimeter while defenders aim to capture intruders. Given the dynamic constraint of the agents, defenders can adjust their actuators to reposition themselves while perceiving the intruders, and the reward function could be the time difference between defenders and intruders to reach the perimeter from the current positions. Modeling this problem in the real-world setting may be challenging, and reinforcement learning can help achieve the goal with active perception.